\documentclass[10pt]{article}

\usepackage{microtype}
\usepackage{graphicx}
\usepackage{subfigure}
\usepackage{booktabs}
\usepackage{hyperref}
\usepackage[preprint]{icml2019}

\usepackage{url}

\usepackage[utf8x]{inputenc}
\usepackage[T1]{fontenc}
\usepackage{titlesec}
\usepackage{amsmath}
\usepackage{amsfonts}
\usepackage{graphicx}
\usepackage[colorinlistoftodos]{todonotes}

\icmltitlerunning{Hamiltonian Monte-Carlo for Orthogonal Matrices}

\begin{document}

\twocolumn[
\icmltitle{Hamiltonian Monte-Carlo for Orthogonal Matrices}

\begin{icmlauthorlist}
\icmlauthor{Viktor Yanush}{jsl,msu}
\icmlauthor{Dmitry Kropotov}{jsl,msu}
\end{icmlauthorlist}

\icmlaffiliation{jsl}{Joint Samsung-HSE lab}
\icmlaffiliation{msu}{Lomonosov Moscow State University}

]
\icmlcorrespondingauthor{Viktor Yanush}{yanushviktor@gmail.com}
\printAffiliationsAndNotice{}

\begin{abstract}
We consider the problem of sampling from posterior distributions for Bayesian models where some parameters are restricted to be orthogonal matrices. Such matrices are sometimes used in neural networks models for reasons of regularization and stabilization of training procedures, and also can parameterize matrices of bounded rank, positive-definite matrices and others. In \citet{byrne2013geodesic} authors have already considered sampling from distributions over manifolds using exact geodesic flows in a scheme similar to Hamiltonian Monte Carlo (HMC). We propose new sampling scheme for a set of orthogonal matrices that is based on the same approach, uses ideas of Riemannian optimization and does not require exact computation of geodesic flows. The method is theoretically justified by proof of symplecticity for the proposed iteration. In experiments we show that the new scheme is comparable or faster in time per iteration and more sample-efficient comparing to conventional HMC with explicit orthogonal parameterization and Geodesic Monte-Carlo. We also provide promising results of Bayesian ensembling for orthogonal neural networks and low-rank matrix factorization.
\end{abstract}

\section{Introduction}

In the paper we consider the problem of sampling from posterior distributions for Bayesian models where some parameters are restricted to be orthogonal matrices. Examples of such models include orthogonal recurrent neural network~\citep{arjovsky2016unitary}, where orthogonal hidden-to-hidden matrix is used to facilitate problems with exploding or vanishing gradients, orthogonal feed-forward neural network~\citep{huang2017orthogonal}, where orthogonal matrices in fully-connected or convolutional layers give additional regularization and stabilize neural network training. Besides, orthogonal matrices can be used in parameterizations for matrices of bounded rank, positive-definite matrices or tensor decompositions. 

It is known that a set of orthogonal matrices forms a Riemannian manifold~\citep{tagare2011notes}. From optimization point of view, using the properties of such manifolds can accelerate optimization procedures in many cases. Examples include optimization w.r.t. low-rank matrices~\citep{vandereycken2013lowrank} and tensors in tensor train format~\citep{steinlechner2015riemannTT}, K-FAC approach for training neural networks of different architectures~\citep{martens2015kfac, martens2018kfac_rnn} and many others. Hence, it is desirable to consider the Riemannian manifold structure of orthogonal matrices within sampling procedures.

One of the major approaches for sampling in Bayesian models is Hamiltonian Monte Carlo (HMC)~\citep{neal2011mcmc}. It can be applied in stochastic setting~\citep{chen2014stochastic} and can consider Riemannian geometry~\citep{ma2015complete}. However, application of these methods for sampling w.r.t. orthogonal matrices requires explicit parameterization of the corresponding manifold. In case of quadratic matrices one possible choice of unambiguous parameterization is given in~\citep{lucas2011lagrangian}. However, it requires computing of matrix exponential which is impractical for large matrices. Another option is to consider parameterization $Q=X(X^TX)^{-1/2}$, where $X$ is arbitrary [possibly rectangular] matrix. This parameterization is non-unique and in practice, as we show in this paper, could lead to slow distribution exploration.  

In this paper we propose a new sampling method for Bayesian models with orthogonal matrices. Here we extend HMC approach using ideas from Riemannian optimization for orthogonal matrices from~\citep{tagare2011notes}. In general outline the main idea of the proposed method is the following. Basic Riemannian HMC approach~\citep{girolami2011riemann} supposes introduction of auxiliary momentum variables that are restricted to lie in tangent space for the current distribution parameters. Given some transformation of these parameters we consider vector transport transformation that maps momentum variables to tangent space for the new parameter values. 
Our method is an extension of previously known Geodesic Monte-Carlo (Geodesic HMC, gHMC) ~\citep{byrne2013geodesic}. The major difference is that step along the geodesic is replaced with retraction step which is much cheaper in terms of computation. 
We prove that the proposed transformation of parameters and momentum variables is symplectic~-- a sufficient condition for correctness of HMC sampling procedure.

We show in experiments that the proposed sampling method is way more sample-efficient than standard HMC with parameterization $Q=X(X^TX)^{-1/2}$. This is true both for non-stochastic and stochastic regimes. We also consider Bayesian orthogonal neural networks for MNIST and CIFAR-10 datasets and show that Bayesian emsembling can improve classification accuracy. For the sake of generality we consider low-rank matrix factorization problem and also show that Bayesian ensembling here can improve prediction quality.

\section{Theory}
In this section we first review the conventional Hamiltonian Monte-Carlo approach and discuss some properties of orthogonal matrices manifold. Then we describe our method called Orthogonal HMC (oHMC) and Orthogonal Stochastic Gradient HMC (oSGHMC) for non-stochastic and stochastic regimes correspondingly.

\subsection{Hamiltonian Monte-Carlo}
Suppose we want to sample from distribution $\pi(\theta) \propto \exp(-U(\theta))$ which we know only up to normalizing constant. For example, we may want to sample from posterior distribution of model parameters given data, i.e. $U(\theta) = -\log p(Data\mid\theta) - \log p(\theta)$. In this situation we can resort to Hamiltonian Monte-Carlo approach that is essentially a Metropolis algorithm with special proposal distribution constructed using Hamiltonian dynamics. Let's consider the following joint distribution of $\theta$ and auxiliary variable $r$:
$$\pi(\theta, r) \propto \exp\left(-U(\theta) - \frac{1}{2}r^T M^{-1}r\right) = \exp\left(-H(\theta, r)\right).$$ 
The function $H(\theta, r)$ is called \emph{Hamiltonian}. Then for sampling from $\pi(\theta)$ it is enough to sample from the joint distribution $\pi(\theta,r)$ and then discard auxiliary variable $r$. As joint distribution factorizes
$$\pi(\theta, r) = \pi(r)\pi(\theta \mid r),$$
we can sample $r \sim \mathcal{N}(0, M)$ and then sample $\theta$ from $\pi(\theta \mid r)$. To do this we simulate the Hamiltonian dynamics:
\begin{align}
    & \frac{d\theta}{dt} = \frac{\partial H}{\partial r} = M^{-1} r, \label{eq:hd1}\\
    & \frac{dr}{dt} = -\frac{\partial H}{\partial \theta} = \nabla_\theta \log \pi(\theta).\label{eq:hd2}
\end{align}
It can be shown that solution of these differential equations preserves the value of Hamiltonian, i.e. $H(\theta(t), r(t))=\mathrm{const}$. Hence we can start from $r(0)=r$ and $\theta(0)$ equaling to previous $\theta$, take $(\theta(t), r(t))$ from some $t$ as new proposal and accept/reject it using standard Metropolis formula. Since $H(\theta(t),r(t))=H(\theta(0),r(0))$ the new point will be accepted with probability $p_\text{accept} = 1$. 

In practice differential equations \eqref{eq:hd1}-\eqref{eq:hd2} cannot be integrated analytically for arbitrary distributions $\pi(\theta)$. However, there exist a special class of numerical integrators for Hamiltonian equations which are called symplectic. Such methods are known to approximately conserve Hamiltonian of the system and produce quite accurate solutions ~\citep{hairer2006long}. The most popular symplectic method for HMC is Leapfrog. One iteration of this method looks as follows:
\begin{align}
    & r^{n+\frac{1}{2}} = r^n + \frac{\varepsilon}{2} \nabla_\theta \log \pi(\theta^n),\label{eq:HMC_1} \\
    & \theta^{n + 1} = \theta^n + \varepsilon M^{-1} r^{n + \frac{1}{2}}, \label{eq:HMC_2} \\
    & r^{n + 1} = r^{n + \frac{1}{2}} + \frac{\varepsilon}{2} \nabla_\theta \log \pi(\theta^{n + 1}).\label{eq:HMC_3}
\end{align}
Since numerical methods introduce some error, we need to calculate Metropolis acceptance probability using old and new Hamiltonian:
$$\rho = \exp(H(\theta_\text{old}, r_\text{old}) - H(\theta_\text{new}, r_\text{new})).$$

\subsection{Stochastic and Riemannian HMC}

If $\pi(\theta)$ is posterior distribution of model parameters $\theta$ given $Data=\{x_i\}_{i=1}^N$, then $\nabla_{\theta}\log\pi(\theta)$ can be written as follows:
$$\nabla_\theta \log \pi(\theta) = \sum_{i=1}^N \nabla_\theta \log p(x_i \mid \theta) + \nabla_\theta \log p(\theta).$$
When $N$ is big enough the sum becomes hard to compute. The straightforward idea is to compute stochastic gradient over a minibatch instead:
$$\nabla_\theta \log \pi(\theta) = \frac{N}{B}\sum_{i=1}^B \nabla_\theta \log p(x_i \mid \theta) + \nabla_\theta \log p(\theta).$$
However, this introduces noise in the Hamiltonian equations and forces to change integration method to counteract this. Moreover, exact calculation of Metropolis acceptance probability becomes intractable since we need to sum log-likelihood over all the objects. This forces to use very small step size to make acceptance probability approximately one. Stochastic algorithms can be found in~\citet{chen2014stochastic}.

Suppose now that we want to sample from distribution $\pi(\theta)$, but $\theta$ lies on a Riemannian manifold. This means that we are given a Riemannian tensor $G(\theta)$ defining the inner product at every point $\theta$:
$$\langle r_1, r_2 \rangle_\theta = \langle G(\theta) r_1, r_2 \rangle.$$
This inner product has direct influence on how distances are computed in the $\theta$ space and this should be taken into account in efficient sampler in order to make larger steps in parameter space. This was done in \citet{girolami2011riemann} and in \citet{ma2015complete}. The main idea is to adapt the Hamiltonian to account for Riemannian structure:
$$H(\theta, r) = U(\theta) + \frac{1}{2}r^T G(\theta) r.$$
Also, in this case the Leapfrog should be replaced with the Generalized Leapfrog.

It's important to note that both conventional HMC and Riemannian HMC treat parameters $\theta$ and $r$ as unconstrained variables. If this is not true, HMC steps could move parameters off the manifold. Let's consider an example, when we have some distribution $\pi(e)\text{, }e \in E$, where $E$ is two-dimensional ellipse with axes $2a$ and $2b$. We can parameterize the points on the ellipse at least in two ways:
\begin{enumerate}
    \item $e = (a \cos t, b \sin t)\text{, where }t \in \mathbb{R}$;
    \item $e = (x, y) \in \mathbb{R}^2$\text{, such that }$\frac{x^2}{a^2} + \frac{y^2}{b^2} = 1$.
\end{enumerate}
In the first case we can apply HMC or Riemannian HMC directly since variable $t$ and auxiliary variable are unconstrained. In this case $G(t) = a^2 \sin^2 t + b^2 \cos^2 t$. In the second case we should have $(x, y)$ lying on the ellipse. As auxiliary variable $r$ is, in a sense, the point velocity it cannot have non-zero component orthogonal to the tangent space. Moreover, even if $r$ lies in the tangent space, modifying $(x,y)$ using equation~\eqref{eq:HMC_2} would move the point off the manifold.

For the case of rectangular orthogonal matrices it is hard to find unambiguous unconstrained parameterization. It means that for sampling on this manifold we need to do iterations that 1) always leave $\theta$ on the manifold and 2) forces $r$ to lie in the tangent space for the current point $\theta$. Hopefully, this can be done using known properties of orthogonal matrix manifold.

\subsection{About manifolds}
\subsubsection{Geodesics, retraction and vector transport}
In this subsection we present some useful definitions and concepts about manifolds. More complete discussion of this topic can be found in \citet{absil2012projection}.

There is a concept of straight line in Euclidean space. On the manifold this concept is extended using notion of \textit{geodesic curve} which is defined as curve $\gamma(t)$ which locally minimizes arc length. Given smooth manifold $\mathcal{M}$ there exists tangent space $T_X \mathcal{M}$ at every point $X \in \mathcal{M}$. Given $X \in \mathcal{M}, U \in T_X \mathcal{M}, t \in \mathbb{R}$ we can define \textit{exponential map} $Exp(X, U)$ as $\gamma(1) = \Gamma(1, X, U)$ where $\Gamma(t, x, u)$ is defined as a curve with the properties:
\begin{equation} \label{eq: retr}
    \begin{split}
    & \Gamma(0, X, U) = X \\
    & \frac{d}{dt}\Gamma(t, X, U)\Big|_{t=0} = U
    \end{split}
\end{equation}
It is known that $Exp(X, tU) = \Gamma(t, X, U)$. In \citet{byrne2013geodesic} authors use exponential map to define the sampling scheme. However, in many cases exact geodesics are quite hard to compute. Any map $R(t, X, U)$ $=$ $R(X, tU)$ which satisfies equations~\eqref{eq: retr} is called \textit{retraction}. Retractions are known to be (at least) first-order approximations of exponential map. Usually, retraction is less expensive computationally than corresponding exponential map.

Along with retraction there is also a concept of \textit{vector transport}. Intuitively, when some point on the manifold moves along a curve its tangent space and all vectors in it are also moving. This process is described via notion of vector transport. Formally, for $X, Y \in \mathcal{M}$ it is defined as a function $T_{X \rightarrow Y}: T_X \mathcal{M} \rightarrow T_Y \mathcal{M}$, which is linear in its argument. A canonical example of such transformation is so-called \textit{parallel transport}. More information can be found in \citet{sato2015new}.

\subsubsection{Stiefel manifold}
In this subsection we present some properties of the manifold of the orthogonal matrices that will be useful later. Many of them can be found in \citet{tagare2011notes}.

Let $\mathcal{V}_p(\mathbb{R}^n) = \{X \in R^{n \times p} \mid X^TX = I\}$ where $n \geq p$ denote the \emph{Stiefel manifold}. There is a convenient representation for the tangent space:
\begin{align*}
    T_X \mathcal{V}_p(\mathbb{R}^n) = \{XA + X_\bot B \mid & A^T = -A \in \mathbb{R}^{p \times p}, \\
    & B \in \mathbb{R}^{(n-p) \times p}, \\
    & [XX_\bot]\text{ is orthogonal}\}
\end{align*}
where $X_\bot$ is an orthogonal complement to a set of orthogonal vectors $X$.

For arbitrary matrix $G\in\mathbb{R}^{n{\times}p}$ it is possible to find projection of this matrix to the tangent space at point $X$ analytically:
$$
    proj_{T_X \mathcal{V}_p(\mathbb{R}^n)} G = G - XG^TX.
$$
Projection is taken with respect to the canonical inner product:
$$\langle Z_1, Z_2 \rangle_X = \operatorname{tr}\left(Z_1^T (I - \frac{1}{2}XX^T) Z_2 \right)$$

For the Stiefel manifold exponential map and parallel transport for $t: (X, U) \rightarrow (X', U')$ are computed simultaneously as follows ~\citep{byrne2013geodesic}: 
\begin{multline}
\left[X', U'\right] = \left[X, U\right]\exp \left( 
t\begin{bmatrix}
    A & -S \\
    I & A
\end{bmatrix}
\right)\times\\
\begin{bmatrix}
    \exp \left(-tA \right) & 0 \\
    0 & \exp \left(-tA\right)
\end{bmatrix},
\label{eq:geodesic_flow}
\end{multline}
where $A = X^T U = -U^T X, S = U^T U$.
Time complexity of that update is at least two matrix exponentials of size $p \times p$ which can be computed in $O(p^3)$ time plus time complexity of matrix multiplication which is $O(np^2)$ in this case. While in the case of big $n$ and small $p$ computation time for matrix exponentials is almost negligible, it is not the case for the square matrices.

Finally, for Stiefel manifold there are many examples of retractions. Most of them can be found in \citet{absil2009optimization}. One particularly important example is based on Cayley transform.
Let $A^T = -A \in \mathbb{R}^{n \times n}, X \in \mathcal{V}_p(\mathbb{R}^n)$. Let
\begin{align}
    & Y(\varepsilon) = \left(I + \frac{\varepsilon}{2}A \right)^{-1}\left(I - \frac{\varepsilon}{2}A\right)X. \label{eq:curve}
\end{align}
Then it is true that:
\begin{align*} 
    & Y(\varepsilon)^T Y(\varepsilon) = I, \\
    & Y(0) = X, \\
    & \frac{d}{d\varepsilon} Y(\varepsilon) \rvert_{\varepsilon = 0} = -AX.
\end{align*}
Thus, if we take $A = -(GX^T - XG^T)$, then $$-AX = G - XG^TX = proj_{T_X \mathcal{V}_p(\mathbb{R}^n)} G.$$
This retraction is useful because it makes our sampling scheme symplectic which is desirable.

\subsection{oHMC}
Suppose we want to sample from distribution $\pi(\theta)$ of orthogonal matrices. The main idea is to try to adapt the Leapfrog integration \eqref{eq:HMC_1}-\eqref{eq:HMC_3} in order to maintain orthogonality constraint for $\theta$ and importantly, preserve symplecticity of update. At first, we note that at every iteration we should have
$\theta \in \mathcal{V}_p(\mathbb{R}^n)$ and $r \in T_\theta
\mathcal{V}_p(\mathbb{R}^n)$. Secondly, we note that if $\theta \in
\mathcal{V}_p(\mathbb{R}^n)$ then Riemannian gradient $\hat{\nabla}_\theta \log \pi(\theta) \in T_\theta \mathcal{V}_p(\mathbb{R}^n)$ is defined
as a particular vector in the tangent space. Given that the update~\eqref{eq:HMC_1} would keep momentum $r$ in the same tangent space. For updating $\theta$ we can use retraction defined in equation~\eqref{eq:curve}. So, the update formula~\eqref{eq:HMC_2} transforms to the following one:
\begin{align}
    & \theta^{n+1} = Q(\theta^{n}, r^{n}, \varepsilon)\theta^{n} \\
    & Q(\theta, r, \varepsilon) =  \left(I - \frac{\varepsilon}{2}(r\theta^T - \theta r^T) \right)^{-1}\left(I + \frac{\varepsilon}{2}(r\theta^T - \theta r^T)\right) \label{eq:inverse}
\end{align}
Yet, that inevitably changes the tangent space and we lose the property that $r \in T_\theta \mathcal{V}_p(\mathbb{R}^n).$ Nonetheless, we can fix this by making vector transport for $r$ to the new tangent space. An orthogonal matrix $X \in \mathcal{V}_p(\mathbb{R}^n)$ is an orthonormal system of $p$ vectors on a $n$-dimensional sphere. The tangent matrix $Z \in T_X \mathcal{V}_p(\mathbb{R}^n)$ is a matrix of tangent vectors which show the direction of rotation. Apparently, if the points on the sphere are rotated in some direction then their velocities should be rotated in the same direction too. So, $\theta$ and $r$ should be updated at the same time to make sure $r \in T_\theta \mathcal{V}_p(\mathbb{R}^n)$. 
Finally we get the following update:
\begin{equation}
\begin{split}
    & \theta^{n+1} = Q(\theta^{n}, r^{n}, \varepsilon)\theta^{n} \\
    & r^{n+1} = Q(\theta^{n}, r^{n}, \varepsilon)r^{n} 
\end{split}
\end{equation}
Combining these update rules we come to the final scheme:
\begin{equation}
\begin{split}
    & r^{n+\frac{1}{2}} = r^n + \frac{\varepsilon}{2} \hat{\nabla}_\theta \log \pi(\theta^n) \\
    & \theta^{n + 1} = Q(\theta^n, r^{n + \frac{1}{2}}, \varepsilon)\theta^n \\
    & r^{n + 1} = Q(\theta^n, r^{n + \frac{1}{2}}, \varepsilon)r^{n + \frac{1}{2}} + \frac{\varepsilon}{2} \hat{\nabla}_\theta \log \pi(\theta^{n + 1})
\end{split}
\end{equation}
This scheme is symmetric (can be reversed in time) and symplectic. The full proof of symplecticity is given in supplementary material. The sketch of the proof is as follows: first we need to compute Jacobian $J = \frac{\partial (\theta^{n+1}, r^{n+1})}{\partial (\theta^n, r^n)}$. This can be done using Woodbury identity several times to rewrite inverse matrices. Having this done, we need to prove that $J^T A J = A$, where $$A = 
\begin{bmatrix}
    0 & I \\
    -I & 0
\end{bmatrix}.$$
This can be done directly by multiplying these matrices in the block fashion.

In conclusion, algorithm works like HMC, but uses new scheme for orthogonal matrices. Leapfrog consists of three steps, so if there are several groups of parameters, constrained or unconstrained, we can run each of Leapfrog steps independently on each group of parameters. The pseudocode is shown in Algorithm~\ref{alg:oHMC}. The inverse matrix in equation~\eqref{eq:inverse} can be computed more efficiently using Woodbury identity. This allows us to reduce the time complexity from $O(n^3)$ to $O(p^3)$. The overall time complexity is $O(p^3 + np^2) = O(np^2)$. This suggests that it is best to use our method with tall matrices of small rank. While asymptotically this method has the same time complexity as Geodesic Monte-Carlo, in practice it is faster and more stable which is shown in the experiments.

\begin{algorithm}[t]
\caption{oHMC}
\label{alg:oHMC}
    \begin{algorithmic}[1]
    \STATE {\bfseries Input:} initial position $\theta^{(1)}$ and step size $\varepsilon$
    \FOR{$t = 1, 2, \dots$}
        \STATE $r^{(t)} \sim \mathcal{N}(0, I)$ \COMMENT{Sample momentum \emph{in the tangent space}}
        \STATE $(\theta_1, r_1) = (\theta^{(t)}, r^{(t)})$
        \FOR{$n = 1 \text{ to } m$}
            \STATE $r_{n + \frac{1}{2}} \gets r_{n} + \frac{\varepsilon}{2} \nabla_\theta \log \pi(\theta_n)$ \COMMENT {Simulate Hamiltonian dynamics}
            \STATE $\theta_{n + 1} \gets Q(\theta_n, r_{n + \frac{1}{2}}, \varepsilon)\theta_n$ \COMMENT{Do retraction}
            \STATE $\hat{r}_{n + \frac{1}{2}}\gets Q(\theta_n, r_{n + \frac{1}{2}}, \varepsilon)r_{n + \frac{1}{2}}$\COMMENT{Do vector transport for $r_{n+\frac{1}{2}}$}
            \STATE $r_{n + 1}\gets \hat{r}_{n + \frac{1}{2}} + \frac{\varepsilon}{2} \nabla_\theta \log \pi(\theta_{n + 1})$
        \ENDFOR
        \STATE $(\hat{\theta}, \hat{r}) = (\theta_m, r_m)$
        \STATE $u \sim U[0, 1]$ \COMMENT{Run rejection}
        \STATE $\rho = \exp(H(\hat{\theta}, \hat{r}) - H(\theta^{(t)}, r^{(t)}))$
        \IF {$u < \min(1, \rho)$} \STATE $\theta^{(t + 1)} = \hat{\theta}$ \ENDIF
    \ENDFOR
    \end{algorithmic}
\end{algorithm}
Stochastic version of oHMC is introduced in the same way as in SGHMC replacing the Leapfrog with our scheme. The pseudocode is shown in Algorithm~\ref{alg:oSGHMC}.
\begin{algorithm}[t]
\caption{Stochastic Gradient oHMC}
\label{alg:oSGHMC}
    \begin{algorithmic}[1]
    \STATE {\bfseries Input:} initial position $\theta^{(1)}$ and step size $\varepsilon$
    \FOR{$t = 1, 2, \dots$}
        \STATE $r^{(t)} \sim \mathcal{N}(0, I)$ \COMMENT{Sample momentum \emph{in the tangent space}}
        \STATE $(\theta_1, r_1) = (\theta^{(t)}, r^{(t)})$
        \FOR{$n = 1 \text{ to } m$}
            \STATE $\theta_{n + 1} \gets Q(\theta_n, r_{n}, \varepsilon)\theta_n$ \COMMENT{Simulate Hamiltonian dynamics}
            \STATE $\hat{r}_n \gets Q(\theta_n, r_{n}, \varepsilon)r_{n}$
            \STATE $\xi \sim \mathcal{N}(0, 2(\alpha - \hat{\beta}))$
            \STATE $r_{n + 1} \gets (1 - \alpha)\hat{r}_{n} + \varepsilon \nabla_\theta \log \pi(\theta_{n + 1}) + \xi$
        \ENDFOR
        \STATE $(\theta^{t + 1}, r^{t + 1}) = (\theta_m, r_m)$\COMMENT{No rejection step}
    \ENDFOR
    \end{algorithmic}
\end{algorithm}

\section{Experiments}
In this section we give experimental results for {oHMC} and {oSGHMC} methods. The first experiments are toy ones intended to show that the proposed method truly works and can sample from the correct distribution.
Also we compare our method with Geodesic HMC ~\citep{byrne2013geodesic} and show its improved time per iteration and numerical stability along with better effective sample size.
Next we apply {oSGHMC} for Bayesian neural networks where parameters in fully-connected and convolutional layers are given by orthogonal matrices. Finally we experiment with Bayesian models consisting of low-rank matrices parameterized by singular value decomposition. Here we consider both simple low-rank matrix factorization model and neural network with low-rank fully connected layers.

\begin{table}[H] 
    \caption{ESS of HMC, Geodesic HMC and oHMC for matrix mixture problem; $n_\text{samples} = 10000.$ Minimum and median are taken with respect to coordinates.}
    \label{tab: ess_qr}
    \vskip 0.15in
    \begin{center}
        \begin{small} \begin{sc}
        \begin{tabular}{l|l|l}
            & \textbf{min} & \textbf{median} \\
            \hline
             HMC & 20.3 & 63.8 \\
             gHMC & 50.2 & 80.4 \\
             oHMC & 103.8 & 139.7
        \end{tabular}
        \end{sc} \end{small}
    \end{center}
    \vskip -0.1in
\end{table}

\subsection{Toy non-stochastic experiment}
\begin{figure} 
\centering
\includegraphics[width=\linewidth]{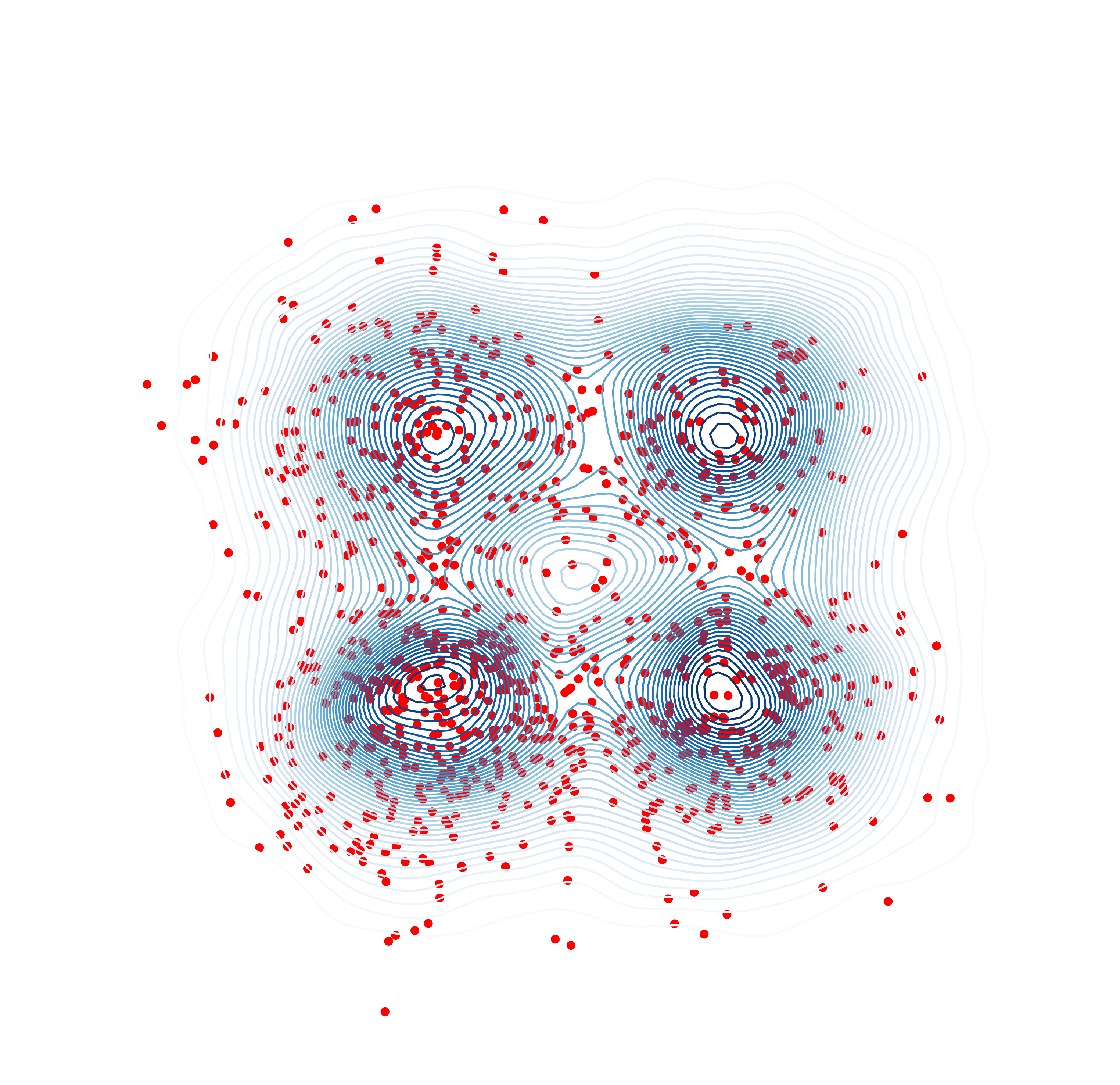}
\caption{Samples from matrix grid in QR parameterization. Blue lines show true density, red dots show samples obtained with oHMC.}\label{fig:qr_matrix_grid}
\end{figure}

Here we consider a mixture of matrix normal distributions as a distribution to sample from. To introduce orthogonal matrices we represent the matrix using QR-decomposition. In other words, the distribution is defined in the following way:
\begin{align*}
\pi(Q, R) = \sum_{i=1}^m \pi_i \mathcal{N}(QR \mid M_i, \sigma^2 I),
\end{align*}
where $Q \in \mathbb{R}^{n \times p}$ is orthogonal and $R \in \mathbb{R}^{p \times p}$ is upper-triangular.
We choose the following parameters: $m = 16, \pi_i = \frac{1}{m}, \sigma = 0.3$. We consider two setups: $n = p = 2$ and $n = 3, p = 2$ to show that our method works both for square and rectangular matrices. Each mode $M_i$ is a matrix where any element $M_{ijk}$ is $1$ or $2$. Initial point for sampling is equal to $M_1$.
We run oHMC for the $n_\text{samples} = 20000$ samples skipping the first $n_\text{burn} = 10000$ samples.Then we plot every tenth sample on some pair of coordinates. The result is shown in figure~\ref{fig:qr_matrix_grid}. Results for square and rectangular matrices are not too different and we show them only for square matrices. To get samples from the true distribution we just sample from the usual matrix normal mixture and then compute QR-decomposition of every sample matrix. As can be seen from the figure, all the modes are covered with sampled points.
We also compare our method with HMC where $Q$ is parameterized as $Q = X(X^T X)^{-1/2}$. This is ambiguous parameterization which can possibly diminish efficiency of the sampler. We compare our method with Geodesic HMC as well. All methods cover the density well enough, however, our method achieves better Effective Sample Size (ESS) which is reported in table~\ref{tab: ess_qr}. Time per iteration of our method is 1.5 times better than of Geodesic HMC and comparable with standard HMC as reported in table ~\ref{tab: time_qr}.

The reason why Geodesic HMC is slower is because of the need to compute matrix exponential twice~\eqref{eq:geodesic_flow}. While time complexity of this operation is asymptotically the same as of matrix inverse, in fact it is more expensive computationally. Another issue with matrix exponential is that it is not quite stable numerically. In this experiment, we could not use larger step size $\varepsilon$ for Geodesic HMC because matrix exponential computation failed. Therefore effective sample size for gHMC is less than for oHMC.

\begin{table}[H] 
    \caption{Time for all 20000 samples for HMC, Geodesic HMC and oHMC for matrix mixture problem; $n_\text{samples} = 10000.$ Measurements are averaged over 5 different runs.}
    \label{tab: time_qr}
    \vskip 0.15in
    \begin{center}
        \begin{small} \begin{sc}
        \begin{tabular}{l|l|l}
            & \textbf{Time (s)} \\
            \hline
             HMC & 13.9 \\
             gHMC & 21.2 \\
             oHMC & 14.7
        \end{tabular}
        \end{sc} \end{small}
    \end{center}
    \vskip -0.1in
\end{table}

\subsection{Toy stochastic experiment}
\begin{figure}[t] 
\centering
\includegraphics[width=\linewidth]{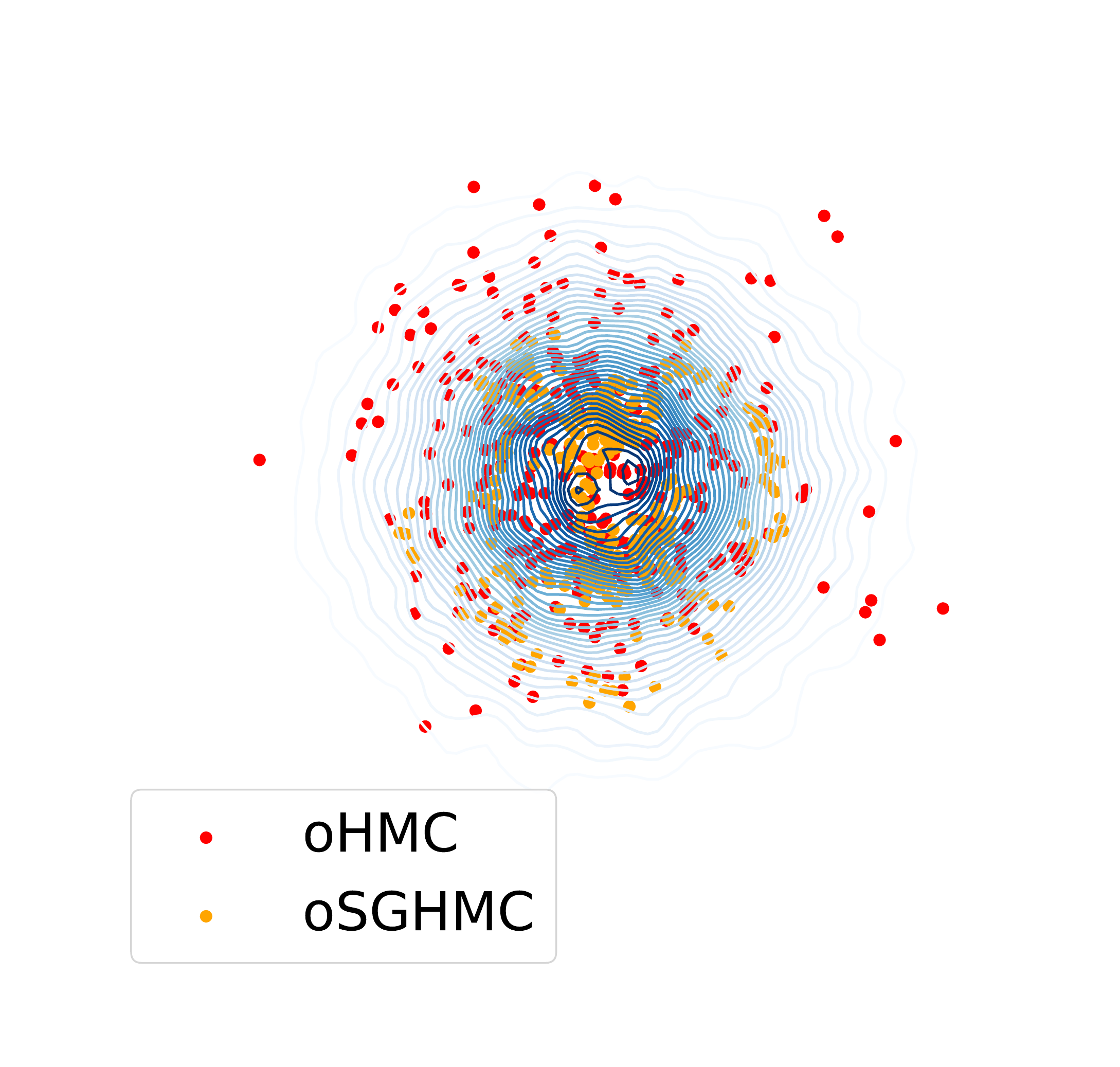}
\caption{Samples from oHMC and oSGHMC matrix grid in QR parameterization. Blue lines show true density distribution. Red dots show oHMC samples and orange dots show oSGHMC samples.}\label{fig:stoch_qr_matrix_grid}
\end{figure}

In this experiment we validate that the stochastic version also works correctly. We take the same model as in the previous experiment, but consider unimodal distribution with $m = 1, \sigma = 1$. We run both oHMC and oSGHMC. For the stochastic method we add artificial noise to the gradients: 
$$\tilde{\nabla} \log \pi(Q, R) = \nabla \log \pi(Q, R) + \mathcal{N}(0, \sigma^2_\text{noise}I),$$
where $\sigma_\text{noise} = 0.1$. As stochastic method needs more iterations to explore the distribution we take $n_\text{samples} = 50000, n_\text{burn} = 10000$. Results are shown in figure~\ref{fig:stoch_qr_matrix_grid}. As we can see, stochastic method still covers the density well and there is no collapse to the mode.
We also compare SGHMC with oSGHMC in terms of ESS. Our method achieves significantly better ESS which shows that our method explores the distribution faster. Results are given in table~\ref{tab: ess_qr_stoch}.
\begin{table}[H] 
    \caption{ESS of SGHMC and oSGHMC in the matrix Gaussian problem; $n_\text{samples} = 30000.$ Minimum and median are taken with respect to coordinates.}
    \label{tab: ess_qr_stoch}
    \vskip 0.15in
    \begin{center}
        \begin{small} \begin{sc}
        \begin{tabular}{l|l|l}
            & \textbf{min} & \textbf{median} \\
            \hline
             HMC & 3.0 & 6.1 \\
             oHMC & 71.7 & 81.2
        \end{tabular}
        \end{sc} \end{small}
    \end{center}
    \vskip -0.1in
\end{table}


\subsection{Orthogonal networks on MNIST and CIFAR-10}
In this experiment we demonstrate that oSGHMC can be applied also to neural networks. For this we train orthogonal VGG-16~\citep{simonyan2014very} on CIFAR-10 and a small orthogonal multilayer perceptron (MLP) on MNIST. By orthogonal network we mean the network where all weight matrices except for the one from the last layer are replaced with orthogonal ones. For convolutional kernel $W \in \mathbb{R}^{o \times i \times w \times h}$ we reshape it to matrix $W \in \mathbb{R}^{o \times iwh}$ and require $W$ to be orthogonal. 
To make sure the performance of sampling and not optimization is measured, we first find good initial point to start sampler from. To do this we optimize the network using oHMC method, where we switch noise to zero. So in fact the method turns into SGD + momentum-like optimization method over orthogonal matrices.

For MNIST dataset we take an orthogonal MLP with the following architecture: 784-300-300-10, where each number shows the number of units on each layer. The first number is the dimensionality of input. The last layer is the usual (non-orthogonal) fully-connected layer. Such choice is motivated by the fact that since orthogonal matrices preserve norm of the input we need to be able to scale them arbitrarily before the softmax. That is, if we make the last fully-connected layer to be orthogonal, then accuracy significantly reduces which proves our intuition.

We run the optimization for 50 epochs and achieve 97.26\% accuracy on the test set. After that we run oSGHMC for $n_\text{samples} = 25000$ samples discarding the first $n_\text{burn} = 5000$. For Bayesian ensemble we take each 1000th sample. Other parameters are: $\varepsilon = 0.00015\text{, batch size} = 256$. The results are given in table~\ref{tab: mnist_ens}.
\begin{table}[t]
    \caption{Test-set accuracy for single network and Bayesian ensemble on MNIST. Standard deviation is taken with respect to 5 independent runs.}
    \label{tab: mnist_ens}
    \vskip 0.15in
    \begin{center}
        \begin{small} \begin{sc}
        \begin{tabular}{l|l|l}
            & \textbf{Single network} & \textbf{Ensemble} \\
            \hline
            Orthogonal & $97.34 \pm 0.08\%$ & $97.87 \pm 0.08\%$ \\
        \end{tabular}
        \end{sc} \end{small}
    \end{center}
    \vskip -0.1in
\end{table}

For CIFAR-10 we run optimization of VGG-16 for 200 epochs and get $92.70 \pm 0.05\%$ accuracy on the test set. In comparison, for the same VGG-16 but with usual weight matrices we need 300 epochs to get the same quality. We want to note that our network has reduced capacity comparing to usual VGG-16 because of orthogonality constraints. Nevertheless, it can achieve almost the same quality. We take learning rate $\varepsilon = 0.1$ and divide it by 10 after 150 epochs. We choose batch size = 128. 

After that we run our sampler for $n_\text{all} = 24000$ samples where we discard first $n_\text{burn} = 4000$ as burn-in and take the last $n_\text{samples} = 20000$. We take each 2000th sample to form an ensemble of 10 networks. Using the ensemble we get $92.85 \pm 0.03\%$ test accuracy. Here we take rather small $\varepsilon = 0.00001$ to make sure the method does not diverge. We gather all results in table~\ref{tab: cifar10_ens}. Results for non-orthogonal networks are taken from~\citet{izmailov2018averaging}.
\begin{table}[t]
    \caption{Test-set accuracy for single network and Bayesian ensemble on CIFAR-10. Standard deviation is taken with respect to 5 independent runs for our method and with respect to 3 runs for results from~\citet{izmailov2018averaging}.}
    \label{tab: cifar10_ens}
    \vskip 0.15in
    \begin{center}
        \begin{small} \begin{sc}
        \begin{tabular}{l|l|l}
            & \textbf{1 network} & \textbf{Ensemble} \\
            \hline
             Non-orthogonal & $93.25 \pm 0.16\%$ & $93.59 \pm 0.16\%$ \\
             Orthogonal & $92.70 \pm 0.05\%$ & $92.85 \pm 0.03\%$ \\
        \end{tabular}
        \end{sc} \end{small}
    \end{center}
    \vskip -0.1in
\end{table}

\subsection{Low-rank matrix factorization}
In this experiment we show how we can use orthogonal parameters to parameterize low-rank matrices. 
The straightforward way to get a low-rank matrix $W \in \mathbb{R}^{m \times n}$ is to make a bottleneck: 
$$W = W_1 W_2\text{, where }W_1 \in \mathbb{R}^{m \times r}, W_2 \in \mathbb{R}^{r \times n}.$$
This factorization is ambiguous since multiplying $W_1$ by a constant and dividing $W_2$ by the same constant leaves $W$ unchanged. This property makes optimization and sampling harder. On the other hand, there exists unique factorization:
\begin{align} 
    & W = U\Sigma V^T, \\
    & U \in \mathbb{R}^{m \times r}, U^T U = I, \\
    & V \in \mathbb{R}^{r \times n}, VV^T = I, \\
    & \sigma_1 \geq \sigma_2 \geq \dots \geq \sigma_r \geq 0.
\end{align}
This is well-known singular value decomposition (SVD). This parameterization is non-redundant and can be much more efficient for optimization and sampling.

We apply these two methods to factorize the rating matrix from the MovieLens-100k dataset. We take the $\text{rank} = 10$. The preprocessing goes as follows: we center all ratings to lie in $[-2, 2]$ where new rating $\hat{r}$ is defined as $\hat{r} = r - 3.$ After that we run optimization to find a good starting point for the sampler. Afterwards, we run our sampler for $n_\text{samples} = 5000$ with $\varepsilon = 0.0001$. We discard the first $n_\text{burn} = 1000$ samples. We take each 500th sample and get an ensemble of 10 factorizations. The results are collected in table~\ref{tab: matrix_factor}.
\begin{table}[t]
    \caption{Results on the MovieLens-100k for low-rank matrix factorization after 20 epochs. The standard deviation is taken w.r.t. 5 runs.}
    \label{tab: matrix_factor}
    \vskip 0.15in
    \begin{center} \begin{small} \begin{sc}
        \begin{tabular}{l|l|l}
             & \textbf{Initial RMSE} & \textbf{Ensemble RMSE} \\
             \hline
             Bottleneck & $1.017 \pm 0.003$ & - \\
             SVD & $1.013 \pm 0.003 $ & $1.001 \pm 0.002$
        \end{tabular}
    \end{sc} \end{small} \end{center}
\end{table}

\subsection{Low-rank MLP on MNIST}
In this experiment we show that low-rank matrix parameterization is applicable for the deep models too. 
Suppose we want to have linear layers with low-rank weight matrices. We may want to do this because this reduces number of layer parameters and allows to make our network smaller hopefully without quality deterioration.

\begin{figure} [t]
\centering
\includegraphics[width=\linewidth]{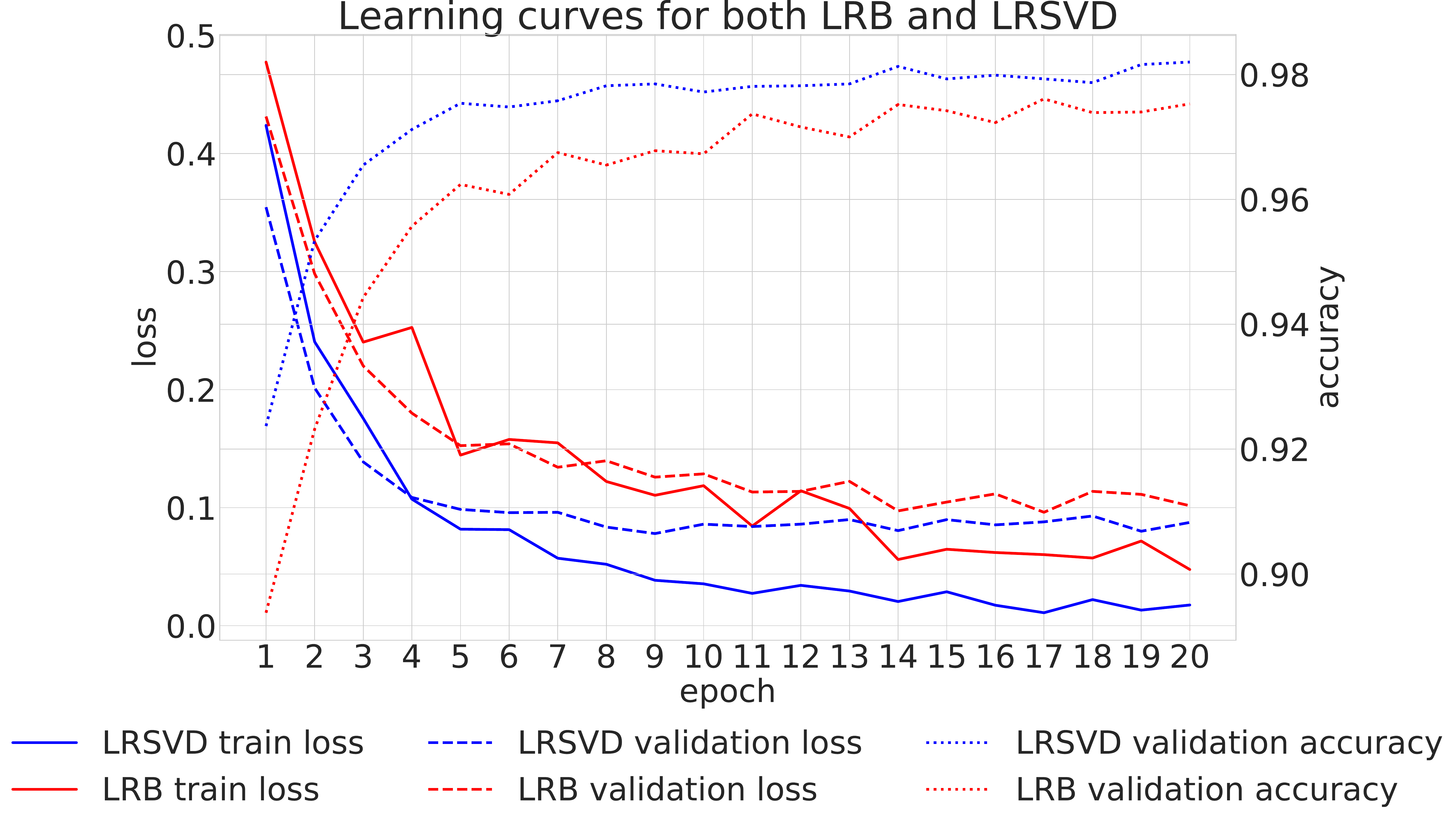}
\caption{Train and validation losses during the training and also validation accuracy for both methods. LRB stands for \textit{Low-rank through Bottleneck}, LRSVD stands for \textit{Low-rank through SVD}.}\label{fig: learning_curve}
\end{figure}

The dataset of choice here is MNIST which is a dataset of 60000 images of size 28x28. The train part is the first 50000 images and the rest is the test part. We construct a multi-layer perceptron (MLP) using only low-rank layers. The architecture is as follows: we have two hidden layers with size of 784 which is quite big, however for the first layer we upper bound the rank by 100 and for the second by 10. The output layer is full-rank. That is, our architecture can be written as 784-784,100-784,10-10,10. We test our method against the same perceptron but in bottleneck parametrization. We train both networks for 20 epochs. The results on the train/test set are given in table~\ref{tab: mnist_low_rank}. The accuracy is quite reasonable for the network without convolutional layers. Moreover, we can see in figure~\ref{fig: learning_curve} that in SVD parameterization the training goes much faster.

We also ran sampling from posterior distribution for our network, however, it did not help much, probably because the accuracy is already quite high for MLP.
\begin{table}[t]
    \caption{Results on MNIST dataset for low-rank MLP after 20 epochs.}
    \label{tab: mnist_low_rank}
    \vskip 0.15in
    \begin{center} \begin{small} \begin{sc}
        \begin{tabular}{l|l|l}
             & \textbf{Train accuracy} & \textbf{Test accuracy} \\
             \hline
             Bottleneck & 98.01\% & 97.53\% \\
             SVD & 99.90\% & 98.25\%
        \end{tabular}
    \end{sc} \end{small} \end{center}
    \vskip -0.1in
\end{table}

\section{Discussion}
In this paper we have considered a new sampling scheme for the manifold of orthogonal matrices. From the results, it is clear that oHMC improves effective sample size comparing to other methods. We have showed that our method is applicable not only for toy problems, but also for the harder tasks such as Bayesian inference for neural networks.

One of the possible concerns about the proposed method is the \textit{irreducibility} property. Roughly speaking, this property means that the Markov Chain which the method constructs can travel through the whole parameter space. For the square matrices it is quite obvious that our method is not irreducible since the retraction step $X \rightarrow R(X, Z)$ preserves $\det X$ but real orthogonal matrices have two possible determinants: $1$ and $-1$. This issue can be resolved by working with unitary matrices as they constitute a connected manifold. 

For the rectangular matrices the answer is not known for us. However, there is the following intuition: for any rectangular orthogonal matrix $X$ we can find an orthogonal complement such that the resulting square orthogonal matrix will have the determinant of desired sign. Therefore, any two rectangular can be connected using our retraction. Although this is not a formal proof, it gives us some hints about the answer to the question about the irreducibility of the method.

\section{Conclusion and Future Work}

In the paper we have presented a new HMC sampler for Bayesian models with orthogonal matrices. Experimental results have shown that new sampling scheme can be much more sample-efficient comparing to conventional HMC with ambiguous parameterization for orthogonal matrices. Using proposed method it becomes possible to make Bayesian ensembling for orthogonal deep neural networks.

Orthogonal matrices can also be used in parameterizations for low-rank matrices, positive-definite matrices and others. It is known that a set of low-rank matrices forms a Riemannian manifold~\citep{vandereycken2013lowrank}. It should be noted that the proposed oHMC and oSGHMC methods do not fully account for this manifold since here matrices $U$ and $V$ in SVD parameterization are allowed to change independently. We consider finding the optimal Riemannian HMC sampler for low-rank matrices for future work. In the same way it is interesting to generalize our sampling scheme for tensors in tensor train format~\citep{steinlechner2015riemannTT}. This family for tensor representation becomes more and more popular and, in particular, can be used within neural networks~\citep{novikov2015tensorizing}.

\bibliographystyle{icml2019}
\bibliography{ohmc_icml2019}

\end{document}